\documentclass[11pt]{article}

\usepackage{amsmath}
\usepackage{amssymb}
\usepackage{graphicx}
\usepackage{hyperref} 
\usepackage{geometry} 
\usepackage{doi}      
\usepackage{adjustbox}
\usepackage{pdflscape}
\usepackage{multirow}
\usepackage{url}
\usepackage{subcaption}

\usepackage{mathtools}
\mathtoolsset{showonlyrefs}

\usepackage[russian, english]{babel}

\usepackage{lscape}

\usepackage{bm}

\title{Finance-Grounded Optimization For Algorithmic Trading}
\author{
  Kasymkhan Khubiyev$*$\\
  Sirius University of Science and Technology, Sirius, Russia \\
  \texttt{kasymkhankhubievnis@gmail.com}
\and
  Mikhail Semenov\\
  Sirius University of Science and Technology, Sirius, Russia \\
  \texttt{semenov.me@talantiuspeh.ru}
\and
    Irina Podlipnova\\
    Sirius University of Science and Technology, Sirius, Russia\\
    Moscow Institute of Physics and Technology, Moscow, Russia\\
    \texttt{podlipnova.iv@talantiuspeh.ru},\\ \texttt{podlipnovas.iv@phystech.edu}\\
\and 
    Dinara Khubieva\\
    Kazan Federal University, Kazan, Russia\\
    \texttt{dinarakhubiyeva@gmail.com}
}

\date{\today}

\begin{document}

\maketitle

\footnotetext{Preprint for the \textit{ICOMP 2025: International Conference on Computational Optimization}
}

\begin{abstract}
Deep neural networks have achieved state-of-the-art performance across a wide range of domains; however, their application to financial decision-making remains challenging due to domain-specific evaluation criteria and interpretability requirements. In quantitative finance, model performance is typically assessed using risk-adjusted and path-dependent metrics that are not naturally aligned with standard training objectives such as mean squared error.

In this work, we introduce a family of financially grounded loss functions derived directly from core quantitative finance metrics, including the Sharpe ratio, profit-and-loss (PnL), and maximum drawdown (MDD). In addition, we propose turnover regularization, a mechanism that explicitly constrains the trading activity induced by model predictions, thereby controlling transaction costs and improving practical deployability.

Empirical results on return prediction tasks demonstrate that models trained with the proposed objectives consistently outperform mean squared error–based baselines when evaluated using algorithmic trading and portfolio-level metrics. These findings suggest that aligning training objectives with financially meaningful performance criteria leads to more effective and economically relevant deep learning models for financial applications.
\end{abstract}

{\centering{\section{Introduction}}}

Deep learning (DL) has become a dominant paradigm across a wide range of application domains, enabling substantial advances in representation learning and predictive modeling. Despite this progress, the application of DL to financial decision-making remains challenging due to the domain’s structural constraints, evaluation criteria, and economic interpretability requirements. Among financial applications, algorithmic trading presents a particularly demanding setting, as models must not only forecast future outcomes but also support decisions that directly affect profit, risk, and capital allocation.

The objective of algorithmic trading is to extract informative signals from heterogeneous data sources—such as market prices, order flow, and unstructured information -- in order to construct strategies that maximize economic performance under uncertainty. While DL models have demonstrated strong predictive capabilities, their training objectives are typically optimized for statistical accuracy rather than financial utility, leading to a misalignment between learning objectives and downstream evaluation metrics.

Recent advances in large language models (LLMs) have further expanded the scope of DL in finance. Techniques such as Chain-of-Thought (CoT) prompting~\cite{wei2023chainofthoughtpromptingelicitsreasoning} and agentic reasoning architectures~\cite{masterman2024landscapeemergingaiagent} have enabled LLMs to perform complex reasoning tasks, including mathematical problem solving. A growing body of work explores the application of LLMs to financial forecasting and decision support~\cite{Zhang2023, Yu2023, lopezlira2023chatgptforecaststockprice}. For instance, Zhang et al.~\cite{Zhang2023} introduced \textit{FinGPT}, a domain-adapted GPT-style model fine-tuned on financial data, demonstrating improved performance over general-purpose LLMs on numerically intensive financial tasks. Yu et al.~\cite{Yu2023} proposed an explainable multimodal forecasting framework based on LLMs, highlighting the challenges of numerical reasoning and addressing them through discrete bin embedding techniques. Lopez and Lira~\cite{lopezlira2023chatgptforecaststockprice} evaluated the ability of \textit{ChatGPT} to forecast stock price movements and constructed a simple trading strategy to illustrate its practical potential. In our previous work~\cite{khubiev-semenov}, we investigated multimodal approaches for stock price prediction, using LLM-based embeddings to encode news flows and integrating them with financial time-series data. The results demonstrated that incorporating unstructured textual information improves forecasting accuracy and outperforms recurrent neural network baselines such as long short-term memory (LSTM) models in multiple settings.

Reinforcement learning (RL) represents another prominent paradigm for sequential decision-making in finance and has been widely adopted in algorithmic trading research. Recent developments, including highly efficient reasoning-oriented models such as DeepSeek R1~\cite{Gou2025}, further emphasize the importance of well-designed reward mechanisms. In financial environments, the effectiveness of RL critically depends on the choice of reward function, which must balance profitability, risk exposure, and trading costs.

Despite methodological diversity, the majority of existing studies rely on standard DL optimization objectives designed for classification or regression tasks. While these objectives -- such as mean squared error (MSE) -- are well-established in general-purpose learning, they are poorly aligned with financial performance criteria. From a financial perspective, minimizing prediction error does not necessarily translate into improved economic outcomes, as practitioners typically evaluate models using risk-adjusted and path-dependent metrics. Consequently, training models with objectives that are decoupled from financial evaluation criteria may lead to suboptimal decision-making behavior.

Several studies have acknowledged this misalignment by incorporating financial metrics into learning objectives or evaluation protocols. For example, in paper~\cite{yanghongyang} profit-and-loss (PnL) was employed as a central component of the reward function in an RL-based trading framework and used the Sharpe ratio to select high-performing strategies over historical periods. The \textit{DianJin-R1} model~\cite{dianjin-r1}, a finance-specialized LLM, focused on improving financial reasoning and interpretability through reinforcement learning with curated financial datasets, including \textit{CFLUE}, \textit{FinQA}, and \textit{CCC}. Although such models demonstrate strong financial reasoning capabilities, they are not explicitly trained to optimize quantitative trading objectives or portfolio-level performance metrics.
The Authors~\cite{Lai2018} propose the alternating direction method of multipliers (ADMM) to handle $\ell_1$-regularization and self-financing constraints simultaneously. They say such approaches enable more aggressive and effective wealth accumulation compared to standard defensive strategies.

Motivated by these limitations, this paper proposes financially grounded loss functions derived directly from fundamental quantitative finance metrics for algorithmic trading and portfolio management. The proposed objectives are designed to be used both as direct training losses for position generation and as components of reward functions in learning-based trading systems.

\textbf{Contributions}

The main contributions of this work are summarized as follows:
\begin{enumerate}
\item We introduce finance-grounded loss functions, including Sharpe ratio–based, maximum drawdown–based, and profit-and-loss–based objectives, that are more suitable for financial time-series modeling than conventional regression losses.
\item We propose a turnover regularization mechanism that implicitly controls trading activity during training, improving strategy stability and economic realism.
\item We demonstrate the applicability of the proposed objectives to portfolio management tasks, highlighting their practical relevance.
\end{enumerate}

The remainder of the paper is organized as follows. Section~2 describes the dataset and provides exploratory analysis. Section~3 presents the research methodology, including evaluation metrics and the proposed loss functions. Section~4 details the experimental setup, model architectures, and trading strategies. Section~5 reports the empirical results, and Section~6 concludes the paper and outlines directions for future work.

{\centering
\section{Data}
}

This section describes the data sources, asset selection criteria, and key characteristics of the datasets used in the experimental evaluation. Given the sensitivity of financial modeling to data quality and temporal consistency, particular attention is paid to data provenance, coverage, and sampling resolution.

We use historical market data obtained from Binance, one of the largest centralized cryptocurrency exchanges (CEXs). Binance provides a public application programming interface (API) that enables access to both high-frequency and aggregated market statistics, making it well suited for empirical studies in algorithmic trading and portfolio analysis.

The backtesting period spans from January~1,~2022 to July~1,~2025. Asset selection was performed by identifying cryptocurrencies that were listed no later than 2021 and remained actively traded (i.e., not delisted) throughout the entire experimental horizon. The year 2021 was intentionally excluded due to extreme market conditions. As shown in Table~\ref{tab:annual_price_change}, the transition from 2021 to 2022 exhibits a median absolute annual price change of $432.42\%$, indicating a highly nonstationary regime that could distort model training and evaluation. Applying these criteria resulted in a universe of $N=61$ cryptocurrencies.

For each selected asset, market data were collected at three temporal resolutions: daily, hourly, and 15-minute intervals. This multi-frequency design enables evaluation of the proposed methods across different trading horizons. The extracted variables include open, high, low, and close prices; base and quote asset trading volumes; taker buy volumes for both base and quote assets; and the total number of executed trades. These features constitute a standard representation of market activity and are widely used in quantitative trading research.

\begin{table}[ht!]
\centering
\begin{tabular}{|c|c|c|c|}
\hline
    Years & BTCUSDT & ETHUSDT & SOLUSDT \\\hline
    2021--2022 & 106.50 & 57.04 & 2.54 \\\hline
    2022--2023 & 16.51 & 7.14 & 81.67 \\\hline
    2023--2024 & 56.77 & 41.47 & 85.23 \\\hline
    2024--2025 & 33.66 & 23.02 & 2.28 \\\hline
\end{tabular}
\caption{Median absolute annual percentage price change of selected highly liquid cryptocurrencies over the study period.}
\label{tab:annual_price_change}
\end{table}

{\centering
\section{Methodology}
}

We study multiple financial data modalities from the perspective of algorithmic trading strategies (alphas)~\cite{Kakushadze2016}. Focusing primarily on candlestick-based market data, we construct trading signals using heuristic approaches, classical machine learning models, and deep neural networks. Our primary objective is to analyze how the choice of training objective—specifically, finance-grounded loss functions—affects the economic performance of the resulting trading strategies.

{\centering\subsection{Problem Statement}}

Consider a universe of $N$ tradable assets observed at discrete time steps $t = 1, \dots, T$. Let $\vec{r}_t \in \mathbb{R}^N$ denote the vector of realized asset returns at time $t$, and let $\vec{x}_t$ represent the corresponding input features derived from market data. A parametric model $f(\cdot;\boldsymbol{\theta})$ maps input features to real-valued scores,
\[
\vec{s}_t = f(\vec{x}_t; \boldsymbol{\theta}),
\]
which are subsequently transformed into portfolio weights through a fixed, differentiable mapping $g(\cdot)$:
\[
\vec{w}_t = g(\vec{s}_t).
\]

The transformation $g(\cdot)$ plays a central role in the trading system, as it enforces portfolio-level constraints and converts model outputs into executable positions. In this work, $g(\cdot)$ is designed to produce \emph{market-neutral} portfolios satisfying
\[
\mathbf{1}^\top \vec{w}_t = 0,
\]
where $\mathbf{1} \in \mathbb{R}^N$ denotes the vector of ones, and to respect a leverage constraint
\[
\|\vec{w}_t\|_2 \leq Q,
\]
with $Q$ denoting the portfolio leverage, typically set to $Q=1$.

Integrating such financial constraints directly into learning-based optimization frameworks often requires specialized algorithmic techniques. For example, Lai et al.~\cite{Lai2018} employed a saddle-point formulation within an ADMM-based solver to enforce self-financing constraints while promoting portfolio sparsity, illustrating the effectiveness of embedding financial structure into iterative optimization procedures.

The learning objective is to optimize the model parameters $\boldsymbol{\theta}$ to improve both return forecasting accuracy and, more importantly, the downstream trading performance induced by the resulting portfolio allocations.

A standard baseline objective for return prediction is the mean squared error (MSE), defined as
\begin{equation}
\mathcal{L}_{\mathrm{MSE}} =
\frac{1}{N}\sum_{i=1}^{N}
\left( \hat{y}_{i,t} - f_i(\boldsymbol{\theta}, \vec{x}_t) \right)^2,
\label{eq:mse_loss}
\end{equation}
where $\hat{y}_{i,t}$ denotes the realized return of asset $i$ at time $t$.

For linear regression models, minimizing~\eqref{eq:mse_loss} yields the best linear unbiased estimator under the Gauss--Markov theorem. While this result does not require normally distributed errors, it assumes homoskedasticity and the absence of serial correlation. These assumptions are routinely violated in financial time series, particularly for asset returns, which exhibit heavy-tailed distributions, volatility clustering, and time-varying conditional variance.

\begin{figure} 
    \centering 
    \includegraphics[width=1.0\linewidth]{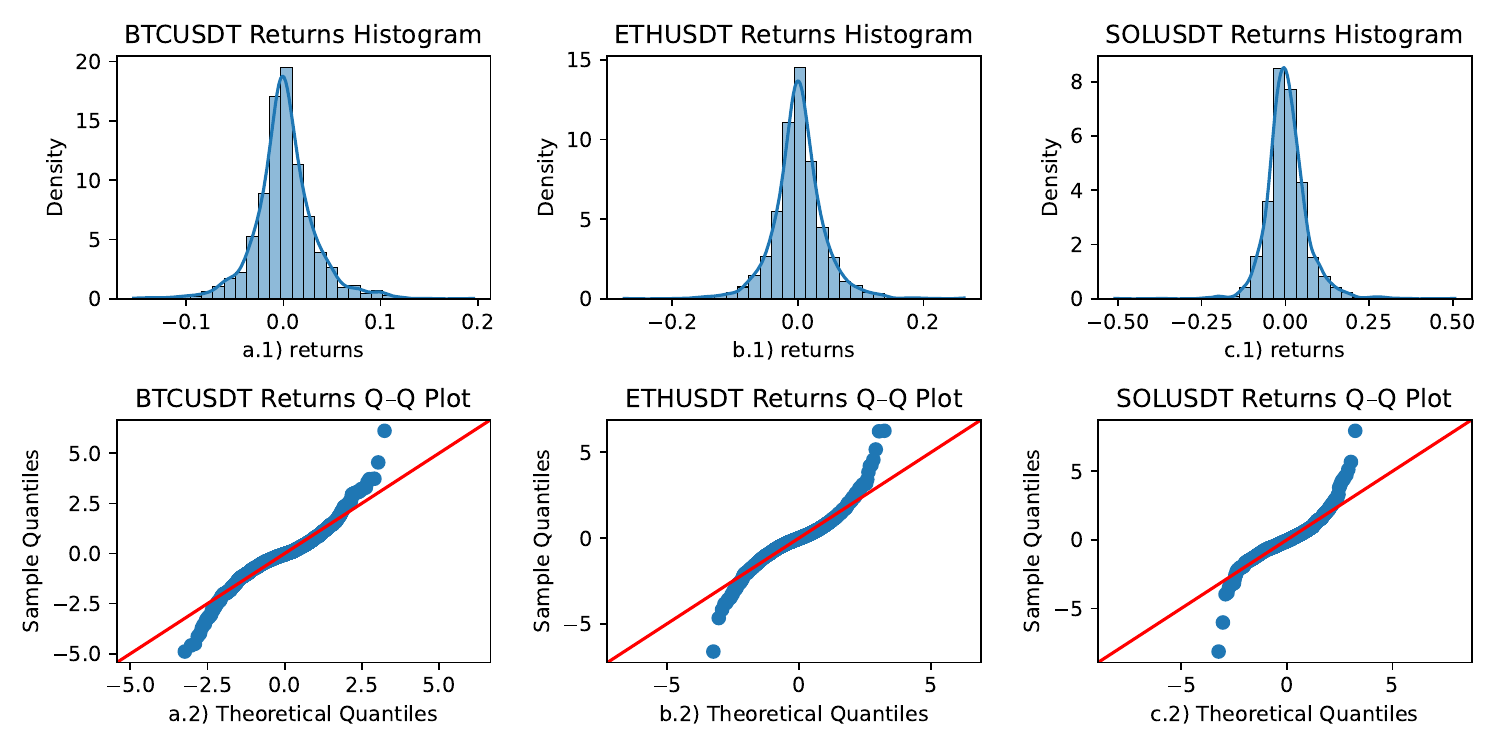} 
    \caption{Return distributions and QQ plots for BTCUSDT, ETHUSDT, and SOLUSDT. Top: histograms of returns with density estimates; bottom: QQ plots against the normal distribution, highlighting heavy-tailed characteristics.} \label{fig:qq_returns} 
\end{figure}

Figure~\ref{fig:qq_returns} illustrates the empirical return distributions of three representative cryptocurrencies—BTCUSDT, ETHUSDT, and SOLUSDT—together with corresponding quantile--quantile (QQ) plots against the normal distribution. The pronounced S-shaped deviations observed in the QQ plots indicate substantial departures from Gaussianity. Table~\ref{tab:shapiro_test_table} reports the results of Shapiro--Wilk and Pearson normality tests, further confirming the non-Gaussian nature of return distributions.

High volatility is a defining characteristic of cryptocurrency markets. While it increases predictive uncertainty, it simultaneously amplifies the economic impact of model errors and highlights the limitations of variance-based regression objectives for optimizing trading performance.

Motivated by these observations and building upon the concept of differentiable financial objectives, we refine standard Sharpe-based loss formulations to address scale insensitivity and numerical stability. Similar to the approach explored in~\cite{Jia2022}, our methodology emphasizes direct optimization of risk-adjusted performance while incorporating variance normalization and auxiliary prediction penalties to improve robustness.

\begin{table}[ht!] 
    \centering 
\begin{tabular}{|l|rr|rr|}
\hline
\multicolumn{1}{|c|}{\multirow{2}{*}{Coin}} & \multicolumn{2}{c|}{Shapiro}       & \multicolumn{2}{c|}{Pearson}            \\ \cline{2-5} 
\multicolumn{1}{|c|}{}                      & \multicolumn{1}{r|}{$W$} & p-value & \multicolumn{1}{r|}{$\chi^2$} & p-value \\ \hline
BTCUSDT & \multicolumn{1}{r|}{0.95} & 1.05e-23  & \multicolumn{1}{r|}{146.19} & 1.80e-32 \\ \hline
ETHUSDT & \multicolumn{1}{r|}{0.94} & 1.39e-25 & \multicolumn{1}{r|}{208.26} & 5.97e-46 \\ \hline
SOLUSDT & \multicolumn{1}{r|}{0.91} & 1.54e-25 & \multicolumn{1}{r|}{335.49} & 1.41e-73 \\ \hline
\end{tabular}
    \caption{Shapiro–Wilk ($W$) and Pearson ($\chi^2$) normality test statistics for daily cryptocurrency return distributions.} 
\label{tab:shapiro_test_table} 
\end{table}

{\centering
\subsection{Evaluation Metrics}
}

Although the collected dataset includes observations at a 15-minute frequency, portfolio rebalancing is performed at a daily frequency. This conservative choice reflects the focus of the study on \emph{medium-frequency} trading strategies, for which positions are adjusted at most once per day. All portfolio rebalance orders are assumed to be executed within the time interval determined by the sampling frequency of the data, ensuring temporal consistency between signal generation and execution.

To benchmark learning-based strategies, we consider several classical heuristic trading strategies commonly used in quantitative finance~\cite{Kakushadze2016}. Specifically, we implement reversion, momentum, mean reversion, and a conservative buy-and-hold (\textit{Buy\&Hold}) strategy, defined as
\begin{equation}
\begin{aligned}
    \text{Reversion}(d)   & = - r(d-1),\\ 
    \text{Momentum}(d)    & = MA(r, w), \\
    \text{Mean Reversion}(d) & = MA(r, w) - r(d),
\end{aligned}
\label{eq:baseline_strategies}
\end{equation}
where $d$ indexes trading days, $MA(\cdot, w)$ denotes a moving average with window size $w$, and $r(d)$ represents the asset return on day $d$, defined as
\begin{equation}
r(d) = \frac{p(d)}{p(d-1)} - 1,
\label{returns_func}
\end{equation}
with $p(d)$ denoting the asset price at day $d$. Returns provide a scale-invariant representation of price dynamics, enabling meaningful comparison across assets.

For each strategy (alpha), we define a \emph{universe} as the set of tradable assets considered at a given time. Each alpha produces a vector of portfolio positions whose dimension equals the number of assets in the corresponding universe. The reversion strategy assumes that recent price movements tend to reverse, momentum exploits trend persistence, and mean reversion trades against deviations from a moving average. The Buy\&Hold strategy allocates capital proportionally across assets and maintains fixed positions throughout the evaluation period.

For machine learning–based strategies, we employ linear regression models. Deep learning baselines include a multilayer perceptron (MLP) and long short-term memory (LSTM) recurrent neural networks. For all models, we evaluate both \emph{single-model} and \emph{ensemble} forecasting settings. In the single-model setting, a model directly predicts a vector of asset scores. In the ensemble setting, separate models are trained per asset and their outputs are aggregated into a unified prediction vector.

Training samples are constructed using data from all three temporal resolutions. Close prices are first transformed into returns using~\eqref{returns_func}, with daily returns serving as target variables. Input features are generated using a sliding window of 20 days: the first 14 days consist of daily returns, followed by 3 days of hourly returns and 3 days of 15-minute returns. This design reflects the assumption that information closer to the execution time benefits from higher temporal resolution, enabling models to capture both long-term trends and short-term market dynamics.

To evaluate and compare the performance of different alphas, we use standard algorithmic trading metrics: the Sharpe ratio, profit and loss (PnL), maximum drawdown (MDD), and turnover. The Sharpe ratio is defined as
\begin{equation}
\text{Sharpe ratio} = \sqrt{H}\,\frac{\mathbb{E}(pnl)}{\sigma(pnl)},
\label{eq:Sharpe_ration}
\end{equation}
where $\mathbb{E}(\cdot)$ and $\sigma(\cdot)$ denote the empirical mean and standard deviation, respectively, $\mathbf{pnl} = (\alpha_1 r_1, \alpha_2 r_2, \ldots, \alpha_M r_M)$ is the vector of per-asset profit-and-loss contributions, $\boldsymbol{\alpha} = (\alpha_1, \ldots, \alpha_M)$ denotes portfolio weights, $\mathbf{r} = (r_1, \ldots, r_M)$ represents realized asset returns, $M$ is the number of assets in the universe, and $H$ is the forecasting horizon length.

The total profit and loss is computed as
\begin{equation}
\text{PnL} = \boldsymbol{\alpha} \mathbf{r} = \sum_{i=1}^{M} pnl_i,
\label{eq:pnl}
\end{equation}
while the maximum drawdown is defined as
\begin{equation}
\text{MaxDrawDown} = -\min_{t}DD_t, \quad DD_t = C_t -\max_{u\leq t}C_u, \quad C_t=\sum_{u\leq t}pnl_u,
\label{eq:maxdrawdown}
\end{equation}
where $C_t$ denotes the cumulative profit-and-loss process.

Finally, portfolio turnover is defined as
\begin{equation}
\text{Turnover} = \sum_{i=1}^{M} \left|\alpha_i(d) - \alpha_i(d-1)\right|,
\label{eq:turnover}
\end{equation}
measuring the total absolute change in portfolio positions between consecutive rebalancing dates.

The Sharpe ratio captures the risk-adjusted consistency of returns, with higher values indicating smoother cumulative PnL trajectories. Maximum drawdown quantifies the worst peak-to-trough loss experienced by a strategy, while PnL reflects the final accumulated profit relative to the initial capital. These metrics are interdependent; for instance, large drawdowns typically result in lower Sharpe ratios due to increased return 

{\centering\subsection{Custom Loss Functions}}
\label{sec:CustomLossFunctions}

For return forecasting, a standard baseline objective is the mean squared error (MSE) loss, which is widely used in regression tasks. However, generic regression objectives are not necessarily aligned with the criteria used to evaluate trading strategies. In particular, minimizing pointwise prediction error does not directly optimize portfolio-level economic performance. To address this objective--metric mismatch, we introduce a set of \emph{finance-grounded} training objectives derived from key trading metrics: the Sharpe ratio (\textit{SharpeLoss}), maximum drawdown (\textit{MDDLoss}), and profit-and-loss (\textit{PnLLoss}).

We define the Sharpe ratio as in Eq.~\eqref{eq:Sharpe_ration} and construct a differentiable Sharpe-based objective with the following modifications. First, we omit the $\sqrt{H}$ scaling factor, which depends only on the forecasting horizon and does not affect the optimizer. Second, to improve numerical stability and gradient behavior, we replace the standard deviation with the variance, thereby avoiding the square-root operation in the denominator. The resulting variance-normalized Sharpe objective is defined as
\begin{equation}
    SharpeLoss = \frac{\mathbb{E}(pnl)}{\mathrm{Var}(pnl)+\varepsilon}.
\label{eq:Sharpeloss}
\end{equation}

During optimization, we minimize $-\text{SharpeLoss}$ in order to encourage higher risk-adjusted performance. Using $\mathrm{Var}(\cdot)$ instead of $\sigma(\cdot)$ yields smoother gradients when $\mathrm{Var}(pnl)$ is small, improving numerical stability in practice.

A detailed convergence analysis of gradient-based optimization for the Sharpe-based objective~\eqref{eq:Sharpeloss} is provided in Appendix~\ref{app:Analysis}.

We next examine the behavior of \textit{SharpeLoss} with respect to the magnitude of the predicted positions. In the considered setting, the model outputs portfolio weights based on historical returns. Since daily returns typically have magnitude on the order of $10^{-2}$, if the model predicts positions $\alpha_i \approx 10^{-2}$, the corresponding profit-and-loss contributions satisfy
\[
pnl_i \approx \alpha_i r_i \approx 10^{-4}.
\]
Consequently, $\mathbb{E}(pnl) \approx 10^{-4}$ and $\sigma(pnl)\approx 10^{-4}$ (or equivalently $\mathrm{Var}(pnl)\approx 10^{-8}$), yielding $\textit{SharpeLoss}\approx 1$. Importantly, if the model instead predicts substantially larger positions, for example $\alpha_i \approx 10^{2}$, both the numerator and denominator scale proportionally, and the resulting Sharpe-based value remains of the same order of magnitude. This demonstrates that Sharpe-based objectives are \emph{scale-invariant} with respect to the magnitude of the predicted positions.

This simple analysis highlights a key limitation of the straightforward \textit{SharpeLoss}: it is largely insensitive to the absolute scale of the generated positions. As a result, the model may converge to degenerate solutions that increase position magnitudes without being penalized by the loss. In practice, such behavior is undesirable due to leverage constraints, transaction costs, and risk management considerations.

\begin{figure}
    \centering
    \includegraphics[width=0.9\linewidth]{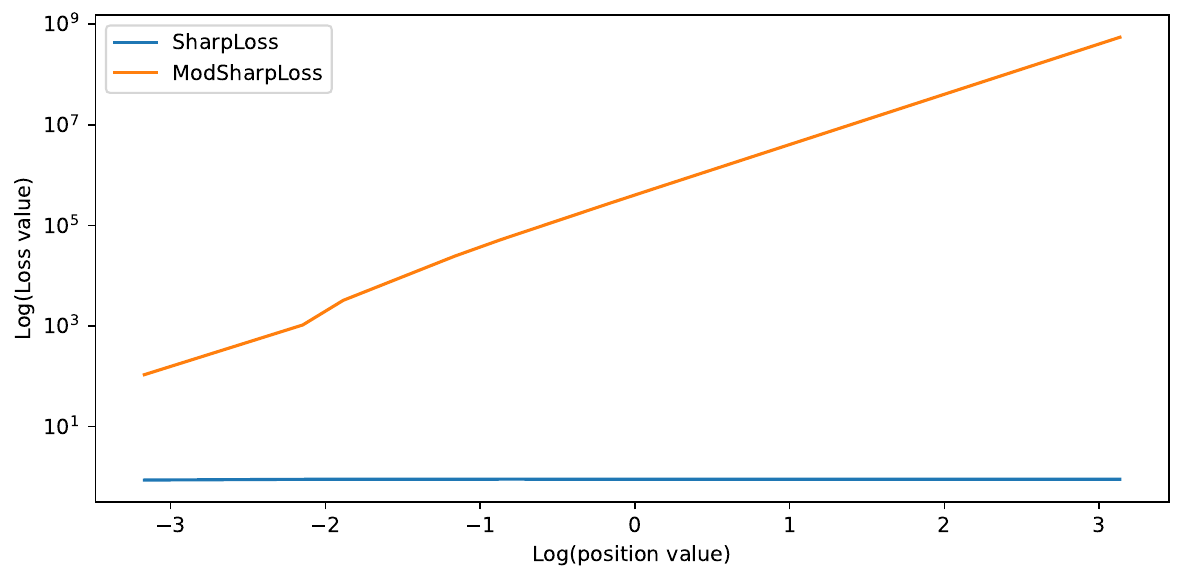}
    \caption{The dependence of the SharpeLoss and ModSharpeLoss values on the magnitude of the generated positions.}
    \label{fig:loss_dependency}
\end{figure}

Figure~\ref{fig:loss_dependency} illustrates this phenomenon empirically by showing loss values as a function of the magnitude of generated positions on a logarithmic scale. The standard \textit{SharpeLoss} remains approximately constant across position magnitudes, whereas the modified objective exhibits an approximately linear dependence. This confirms the scale-insensitivity issue and motivates the proposed modification.

To address this limitation, we construct a loss function that better reflects the task requirements and denote it as the \emph{modified Sharpe loss}, abbreviated as \textit{ModSharpeLoss} or \textit{MS}. Specifically, we augment the Sharpe-based objective with an additional factor that penalizes deviations from the ground-truth return, thereby introducing sensitivity to prediction quality and improving optimization stability.

Assuming the loss function is defined as
\begin{equation}
    ModSharpeLoss = \mathbb{E}[\alpha-r]\frac{\mathbb{E}(pnl)}{\mathrm{Var}(pnl)+\varepsilon},
\label{eq:ModSharpeloss}
\end{equation}

an equivalent formulation can be expressed as
\begin{equation}
    MS(\theta, x) = \mathbb{E}[f(x, \theta) - y]\,S(\theta),
    \label{eq:mod_sharpe_loss}
\end{equation}
where $S(\theta)$ denotes the Sharpe-based objective in~\eqref{eq:Sharpeloss}.

Since $\nabla\mathbb{E}[f - y] = \mathbb{E}[\nabla f]$, the corresponding gradient takes the form
\begin{equation}
    \nabla MS = \nabla\bigl[\mathbb{E}[f - y]\, S\bigr] 
    = S\,\nabla\mathbb{E}f + \mathbb{E}[f - y]\,\nabla S. 
\end{equation}

The modified Sharpe objective~\eqref{eq:mod_sharpe_loss} augments the Sharpe term with an explicit prediction-quality penalty. While this modification alters the optimization landscape, it preserves the smoothness and boundedness properties required for the convergence analysis presented earlier. Consequently, standard gradient-based methods still converge to first-order stationary points of the modified objective.

Importantly, this formulation can be interpreted as a soft regularization mechanism that trades off pure risk-adjusted performance against prediction accuracy.
It discourages degenerate solutions that achieve high Sharpe values through unstable or poorly calibrated position outputs. Similarly, the MS formulation regularizes Sharpe optimization by penalizing strategies that attain high risk-adjusted returns at the expense of predictive reliability.

However, the deviation penalty in~\eqref{eq:mod_sharpe_loss} is sign-unstable: if $\mathbb{E}[\alpha-r]$ becomes negative, it may reverse the effective optimization direction of the Sharpe-like ratio and lead to unstable learning dynamics. To address this issue, we introduce two alternative variants based on norm-type expectation penalties:
\begin{equation}
    ModSharpeAbsLoss = \mathbb{E}[|\alpha-r|]\frac{\mathbb{E}(pnl)}{\mathrm{Var}(pnl)+\varepsilon},
    \label{eq:mod_sharpe_abs_loss}
\end{equation}
and
\begin{equation}
    ModSharpeSquaredLoss = \mathbb{E}[(\alpha-r)^2]\frac{\mathbb{E}(pnl)}{\mathrm{Var}(pnl)+\varepsilon}.
    \label{eq:mod_sharpe_squared_loss}
\end{equation}

We additionally implement several finance-oriented objectives: profit-and-loss (\textit{PnLLoss}) in~\eqref{eq:pnl_loss}, a composite risk-adjusted objective (\textit{RiskAdjLoss}), and maximum drawdown (\textit{MDDLoss}) in~\eqref{eq:mddloss}. As a regression baseline, we use the PyTorch implementation of mean squared error (\textit{MSELoss}). All objectives are formulated as minimization problems: \textit{PnLLoss} is defined with a negative sign because the underlying goal is to maximize profit, while \textit{MDDLoss} explicitly penalizes large drawdowns.

\begin{equation}
    PnLLoss = -\alpha r,
    \label{eq:pnl_loss}
\end{equation}

\begin{equation}
    RiskAdjLoss = -\mathbb{E}(pnl) + \lambda \times DrawDown + \gamma \times (\alpha - r)^2,
\end{equation}
where $\lambda$ and $\gamma$ are the drawdown penalty and position-regularization coefficients, respectively.

\begin{equation}
    MDDLoss = -\min_t DD_t.
    \label{eq:mddloss}
\end{equation}
To improve numerical stability and convergence behavior, we further consider two smooth variants of \textit{MDDLoss}, based on logarithmic transformation and soft-min approximation. 
In particular, the minimum operator for the soft version is approximated as
\begin{equation}
    \min_t z_t \approx -\tau \log\sum_t\exp{(-z_t/\tau)},
\end{equation}
where $\tau>0$ controls the smoothness of the approximation.

Although the objectives above are directly motivated by financial performance criteria, they are generally non-convex and may introduce optimization challenges. To complement these losses, we also consider a set of \emph{convex risk-based} objectives commonly used in portfolio theory, which are naturally expressed as minimization problems. These include the mean--variance loss~\eqref{eq:mean_variance_loss}, mean-CVaR loss~\eqref{eq:mean_cvar_loss}, and entropic risk loss~\eqref{eq:entropic_risk_loss}:
\begin{equation}
    MVLoss = \lambda \cdot \mathrm{Var}(pnl_\theta) - \mathbb{E}[pnl_\theta],
    \label{eq:mean_variance_loss}
\end{equation}

\begin{equation}
    MCVaRLoss = \eta + \frac{1}{1-\alpha}\mathbb{E}[(-pnl_\theta-\eta)_{+}] - \lambda \cdot \mathbb{E}[pnl_\theta],
    \label{eq:mean_cvar_loss}
\end{equation}

\begin{equation}
    ERLoss = \frac{1}{\gamma}\log\mathbb{E}[\exp{(-\gamma \cdot pnl_\theta)}],
    \label{eq:entropic_risk_loss}
\end{equation}
where $\lambda$, $\gamma$, and $\alpha$ are hyperparameters, and $\eta$ and $\theta$ denote optimization variables.

{\centering\subsection{Turnover Regularization}}

We further address the practical requirement of controlling portfolio turnover by introducing a custom turnover regularization penalty. In learning-based trading strategies, unconstrained optimization may converge to nearly static position vectors over time, resulting in near-zero turnover and behavior resembling a buy-and-hold policy. While such solutions minimize transaction costs, they fail to exploit short-term market dynamics and are undesirable in active trading settings.

To explicitly regulate trading activity, we propose a \emph{band turnover regularization} that penalizes deviations of the average portfolio turnover from a predefined admissible range:
\begin{equation}
    \text{TvrReg}
    = \lambda \cdot \bigl(\max(0, \text{tvr} - tb) + \max(0, bb - \text{tvr})\bigr),
    \label{eq:tvrreg}
\end{equation}
where $\text{tvr}$ denotes the average turnover computed as in~\eqref{eq:turnover}, $tb$ and $bb$ represent the upper and lower turnover bounds, respectively, and $\lambda$ controls the penalty magnitude. This formulation introduces a soft constraint: no penalty is applied when turnover lies within the interval $[bb, tb]$, while linear penalties are incurred for excessive trading activity ($\text{tvr} > tb$) or insufficient portfolio reallocation ($\text{tvr} < bb$). Consequently, the model is encouraged to maintain turnover within an economically meaningful range, balancing responsiveness to market signals and transaction cost efficiency.

This approach contrasts with classical turnover regularization commonly employed in portfolio optimization, which directly penalizes trading activity via a linear term $\lambda \cdot \text{tvr}$. While the classical formulation monotonically discourages turnover and often drives learning-based strategies toward overly conservative, near-static portfolios, the proposed band regularizer enforces controlled trading behavior without collapsing to trivial low-turnover solutions. By explicitly specifying acceptable turnover levels, the band-based penalty improves interpretability and operational realism.

To assess the impact of turnover control, we apply \textit{TvrReg} in combination with \textit{MSELoss}, \textit{SharpeLoss}, and \textit{ModSharpeLoss} when training LSTM-based models. In the experiments, we set $\lambda = 1.0$, $tb = 1.0$, and $bb = 0.3$ for turnover regularization, and $\lambda = 0.3$ and $\gamma = 0.01$ for the \textit{RiskAdjLoss}.

The turnover bounds are motivated by the daily rebalancing framework and typical portfolio trading constraints. The upper bound $tb = 1.0$ corresponds to a full portfolio turnover per rebalancing period, preventing unrealistically high trading intensity exceeding the portfolio notional. The lower bound $bb = 0.3$ enforces a minimum reallocation level of $30\%$, ensuring sufficient portfolio adaptation to short-term market signals and avoiding degenerate near-static strategies.

The band turnover regularization induces a piecewise-linear convex penalty landscape with a flat region inside the admissible interval $[bb, tb]$ and linear growth outside the bounds. As a result, optimization remains unconstrained when turnover stays within the target range, while deviations are softly penalized. Increasing the upper bound $tb$ permits more aggressive trading and faster adaptation at the cost of higher transaction intensity, whereas decreasing $tb$ enforces smoother portfolio evolution. Similarly, increasing the lower bound $bb$ promotes more frequent rebalancing, while smaller values allow more conservative strategies.

In practice, the performance of the band regularizer is robust within economically meaningful ranges of $(bb, tb)$. Following common turnover constraints in portfolio optimization, we set $tb$ close to one portfolio notional per rebalancing period and choose $bb$ in the range of $0.1$--$0.3$ to prevent degenerate low-turnover solutions. Extremely narrow bands overly restrict trading flexibility, whereas excessively wide bands reduce the regularizer to a negligible effect. Accordingly, $(bb, tb)$ are selected based on operational considerations and validated through limited sensitivity analysis.

{\centering
\section{Experiment}
}

We conduct a comprehensive experimental evaluation to assess the impact of finance-grounded training objectives on alpha construction and portfolio optimization. All experiments are performed using historical market data from Binance, as described in Section~2.

\paragraph{Baselines and predictive models.}
We begin with three classical heuristic alphas: reversion, momentum, and mean reversion, which serve as non-learning baselines. As machine learning baselines, we implement linear regression (LinReg), random forest (RF), and XGBoost models to predict asset returns. These models are trained using default configurations without hyperparameter grid search, reflecting standard off-the-shelf usage and providing a reference for comparison with deep learning approaches.

For deep learning–based alphas, we employ a multilayer perceptron (MLP) and a long short-term memory (LSTM) network. The MLP consists of three fully connected layers with ReLU activations, while the LSTM follows a standard recurrent architecture with a linear output projection layer. As a baseline objective, we use mean squared error (\textit{MSELoss}). We further evaluate the proposed finance-grounded objectives, including \textit{PnLLoss}, \textit{MDDLoss}, logarithmic and soft \textit{MDDLoss}, \textit{SharpeLoss}, \textit{RiskAdjLoss}, \textit{ERLoss}, \textit{MVloss}, \textit{MCVaRLoss}, and \textit{ModSharpeLoss}, along with their combinations with classical turnover regularization and the proposed band turnover regularizer.

\paragraph{Prediction protocol and portfolio construction.}
We adopt a point-wise prediction framework in which models generate return forecasts for each asset at every trading day prior to portfolio rebalancing. The predicted returns are aggregated into position vectors and subsequently transformed using the differentiable mapping described in Section~3 to enforce market-neutrality and leverage constraints. Specifically, predicted scores are neutralized to ensure zero net exposure and scaled to satisfy the portfolio norm constraint.

Importantly, we do not apply any post-hoc smoothing, clipping, or turnover-reduction techniques after position generation. This design choice allows us to directly evaluate the intrinsic ability of each model and training objective to produce economically meaningful trading signals and stable portfolio dynamics.

\paragraph{Optimization details.}
All deep learning models are trained using the AdamW optimizer. We employ a constant learning rate of $10^{-5}$ for MLP models and $10^{-3}$ for LSTM models, which were found to provide stable and robust convergence across all experiments. No learning rate scheduling or additional regularization beyond the specified loss formulations is applied.

\paragraph{Portfolio optimization experiments.}
In addition to single-alpha construction, we study deep learning approaches for combining multiple trading signals. To this end, we construct a set of 20 low-correlated alphas using multimodal financial data, including candlestick features, order and trade statistics, and limit order book information. As a baseline portfolio, we consider an equally weighted combination of alphas.

We train an LSTM model to generate alpha-combination weights and investigate two portfolio-construction variants:

\begin{itemize}
    \item \textbf{Single-weight generation}: given $L$ alphas, the model outputs a vector of $L$ weights at each time step, where each scalar weight uniformly scales the corresponding alpha position vector.
    \item \textbf{Point-wise weight generation}: given $L$ alphas and $M$ assets, the model outputs an $L \times M$ weight matrix at each time step, enabling asset-specific weighting of alpha positions. The final portfolio position vector is obtained by summing element-wise products between alpha positions and predicted weights.
\end{itemize}

\paragraph{Implementation.}
All deep learning models are implemented using the PyTorch framework and trained on an NVIDIA V100 GPU.

{\centering
\section{Results and Discussion}}

{\centering\subsection{Loss Functions}}

We use LSTM models as the primary deep learning baseline for constructing algorithmic trading strategies (alphas). We first evaluate the impact of different training objectives, including the proposed finance-grounded losses, and analyze the role of turnover regularization~\eqref{eq:tvrreg} in shaping trading performance.

\begin{figure}[ht!]
    \centering
    \includegraphics[width=1.0\textwidth]{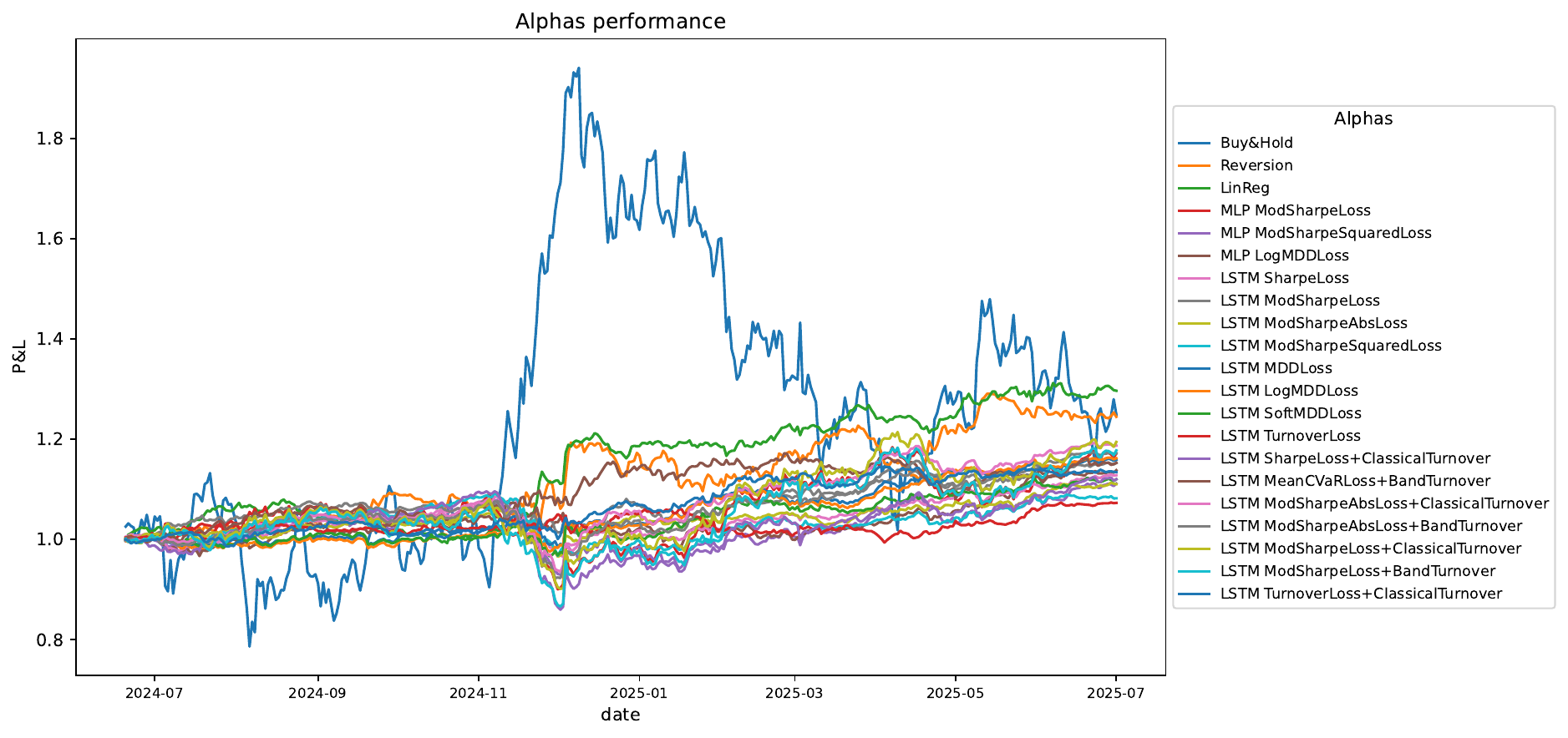} 
    \caption{Cumulative profit-and-loss (PnL) trajectories of the top 15 alphas ranked by Sharpe ratio over the test interval, together with the \textit{Buy\&Hold} benchmark. 
Finance-grounded loss functions produce consistently smoother and higher-performing strategies compared to the passive baseline.}

    \label{fig:alphas_result_test} 
\end{figure}

For clarity, models trained with turnover regularization are denoted using the \textit{TvrReg} suffix. Figure~\ref{fig:alphas_result_test} displays cumulative profit-and-loss (PnL) trajectories of the top 15 alphas ranked by Sharpe ratio over the test interval, providing a visual comparison of the strongest-performing strategies. Table~\ref{tab:alphas_result_test} reports the corresponding quantitative evaluation metrics for top $15$ evaluated models. The whole table containing evaluation metrics for all observed models and losses is presented in the Appendix

Overall, LSTM-based strategies optimized with finance-grounded objectives substantially outperform classical heuristic alphas, machine learning baselines, and the conservative \textit{Buy\&Hold} strategy. In contrast, models trained using standard regression objectives such as \textit{MSELoss} consistently exhibit negative Sharpe ratios and poor economic performance, confirming that minimizing pointwise prediction error is misaligned with optimizing trading profitability.

Among the considered objectives, logarithmic maximum drawdown loss (\textit{LogMDDLoss}) yields the most robust and stable performance. The corresponding LSTM model achieves the lowest maximum drawdown while maintaining one of the highest cumulative PnL values and Sharpe ratios on the test interval. The closely related \textit{MDDLoss} objective exhibits similar behavior, indicating that explicitly incorporating downside risk control into the optimization process leads to superior risk-adjusted returns.

Modified Sharpe-based objectives further improve performance relative to the baseline \textit{SharpeLoss}. By addressing the scale invariance and instability of the standard Sharpe formulation, \textit{ModSharpeLoss} and its variants produce consistently higher Sharpe ratios and smoother PnL trajectories. These results validate the proposed modification and highlight the importance of coupling risk-adjusted optimization with prediction-quality regularization.

An apparent exception is the linear regression baseline, which attains a high Sharpe ratio on the test interval. However, this performance is accompanied by extremely high turnover and pronounced instability over the full backtesting horizon. In particular, LinReg experiences some of the largest drawdowns observed among all evaluated strategies, indicating strong regime sensitivity and limited robustness. The elevated Sharpe ratio observed on the test subset therefore reflects occasional favorable market conditions rather than consistent risk-adjusted performance.

Turnover regularization provides a substantial additional benefit across most learning-based strategies. Both classical and band-based penalties improve economic performance by constraining excessive trading activity and preventing convergence to degenerate near-static portfolios. Notably, the proposed band turnover regularization maintains strong Sharpe ratios while avoiding the over-suppression of trading dynamics often induced by linear turnover penalties. This demonstrates that explicitly controlling turnover within economically meaningful bounds is essential for realistic deployment.

Finally, the constructed alphas exhibit low pairwise correlations. This diversification property indicates that the learned strategies capture complementary market dynamics and are well suited for portfolio-level combination and optimization, which we investigate in the following subsection.

\begin{table}[ht!]
\centering
\small
\begin{tabular}{|l|r|r|r|r|}
\hline
\textbf{Strategy} & \textbf{Turnover} & \textbf{Max Drawdown} & \textbf{PnL (\%)} & \textbf{Sharpe $\uparrow$} \\\hline

LSTM LogMDDLoss  & 0.17 & -0.0637 & 16.17 & \textbf{1.7573} \\\hline
LSTM MDDLoss    & 0.18 & -0.0671 & 15.87 & 1.6914 \\\hline
LSTM ModSharpeAbsLoss + ClassicalTvr & 0.03 & -0.1453 & 18.72 & 1.5551 \\\hline
LSTM ModSharpeLoss & 0.12 & -0.0670 & 15.28 & 1.5283 \\\hline
LSTM TurnoverLoss + ClassicalTvr & 0.04 & -0.0405 & 13.42 & 1.4989 \\\hline
LSTM ModSharpeAbsLoss + BandTvr & 0.04 & -0.1529 & 17.15 & 1.4711 \\\hline
LSTM SharpeLoss & 0.10 & -0.1198 & 12.80 & 1.3628 \\\hline
MLP LogMDDLoss & 0.24 & -0.0919 & 13.78 & 1.2882 \\\hline
LinReg & 1.43 & -0.0870 & 29.64 & 2.0963 \\\hline
Reversion & 0.12 & -0.1083 & 24.45 & 1.2418 \\\hline

XGBoost & 1.42 & -0.0985 & 25.93 & 0.7181 \\\hline
Mean Reversion & 0.76 & -0.2407 & 14.92 & 0.6480 \\\hline
Buy\&Hold & 0.04 & -0.9226 & 24.90 & 0.3072 \\\hline

LSTM MSELoss & 0.22 & -0.2004 & -6.03 & -0.4564 \\\hline
MLP MSELoss & 0.59 & -0.1131 & -6.32 & -0.5924 \\\hline
Random Forest & 0.94 & -0.1074 & -7.30 & -0.7808 \\\hline

\end{tabular}
\caption{\textbf{Performance of algorithmic trading strategies (alphas) over the test interval, sorted by Sharpe ratio.} 
Each row reports portfolio turnover, maximum drawdown, cumulative profit-and-loss (PnL), and Sharpe ratio for heuristic strategies, classical machine learning baselines, and deep learning models trained with different loss functions. 
Finance-grounded objectives—particularly \textit{LogMDDLoss} and modified Sharpe-based losses—consistently yield superior risk-adjusted performance and reduced drawdowns compared to standard regression objectives. 
Although linear regression attains a high Sharpe ratio, it exhibits extremely high turnover and substantial instability, indicating limited robustness. 
Models trained with \textit{MSELoss} perform poorly across all economic metrics, confirming the misalignment between prediction error minimization and trading profitability. 
Suffixes \textit{ClassicalTvr} and \textit{BandTvr} denote classical and band-based turnover regularization, respectively.}
\label{tab:alphas_result_test}
\end{table}

{\centering\subsection{Portfolio Optimization}}

\begin{figure}[ht!]
    \centering
    \includegraphics[width=0.9\textwidth]{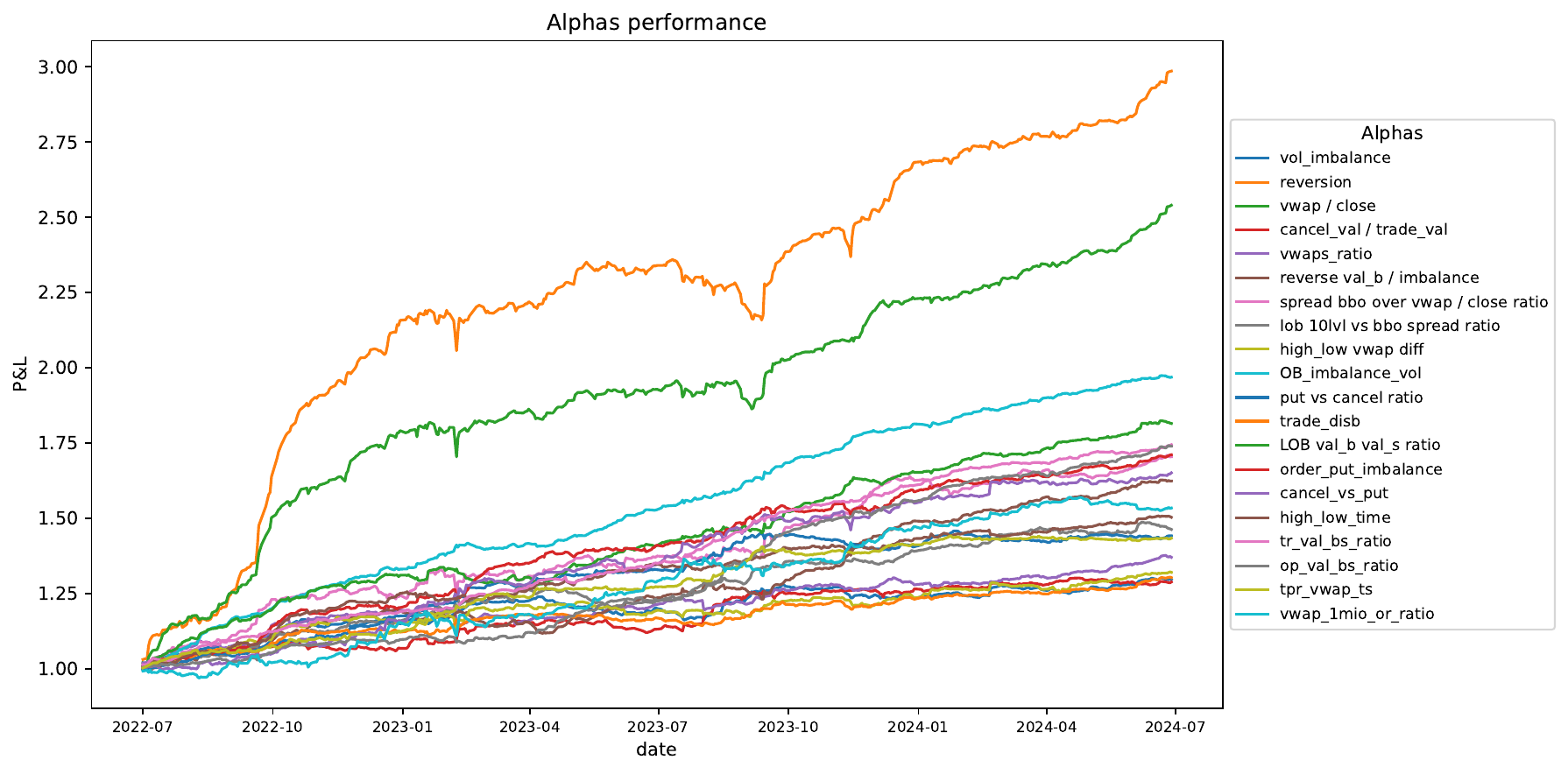} 
    \caption{Alphas performance.} 
    \label{fig:alphas} 
\end{figure}

\begin{figure}[ht!]
    \centering
    \includegraphics[width=0.9\textwidth]{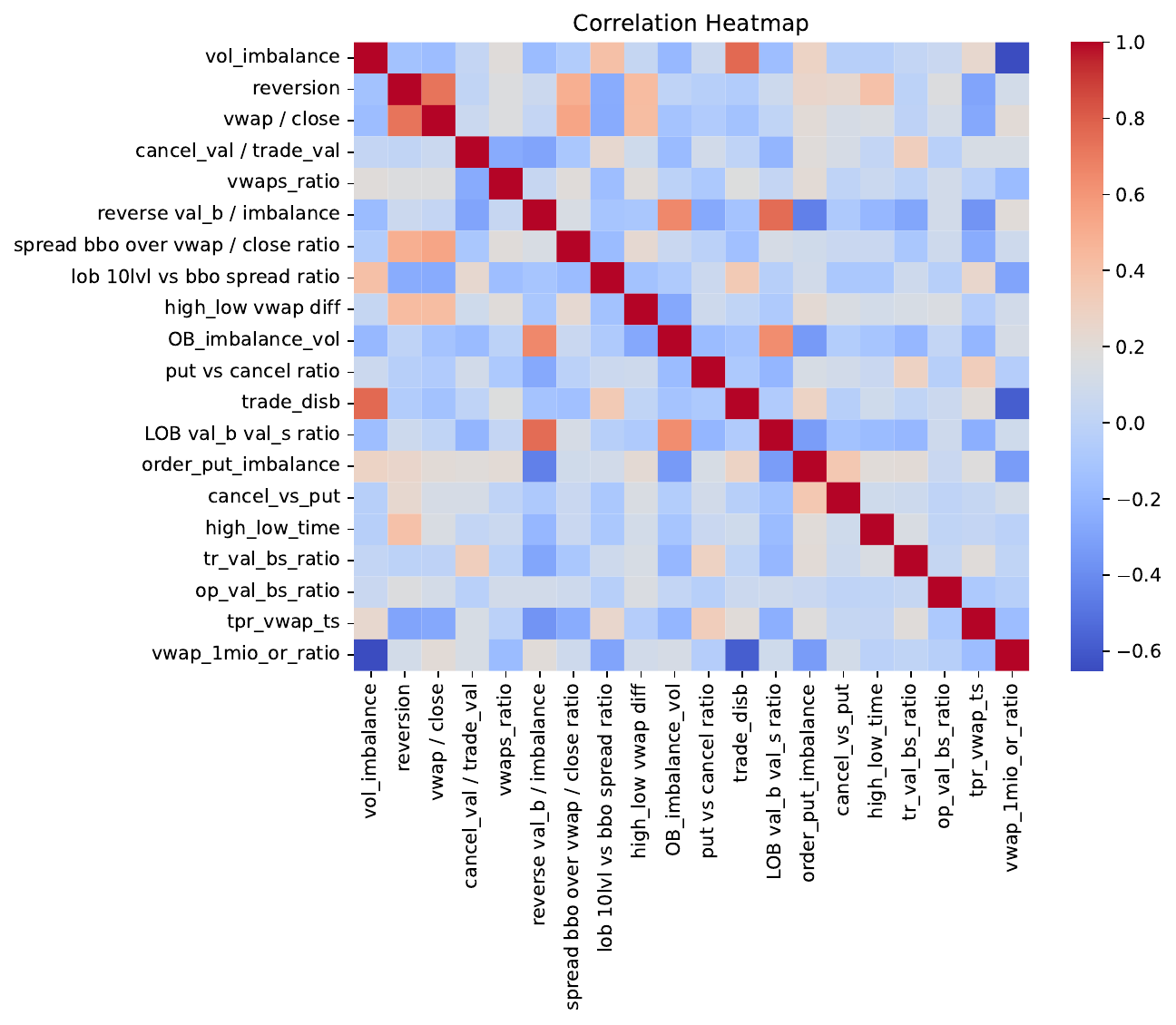} 
    \caption{Alphas correlation heatmap.} 
    \label{fig:classical_alphas_corr_map} 
\end{figure}

To evaluate deep learning–based portfolio optimization techniques, we construct a set of 20 low-correlated alphas using multimodal financial data, including candlestick features, order and trade statistics, and limit order book information. The pairwise correlation structure of the resulting alphas is shown in Figure~\ref{fig:classical_alphas_corr_map}, confirming substantial diversification across trading signals. Figure~\ref{fig:alphas} presents cumulative PnL trajectories of the individual alphas, while Table~\ref{tab:classical_alphas} reports their corresponding performance metrics.

\begin{figure}[ht!]
    \centering
    \includegraphics[width=0.9\textwidth]{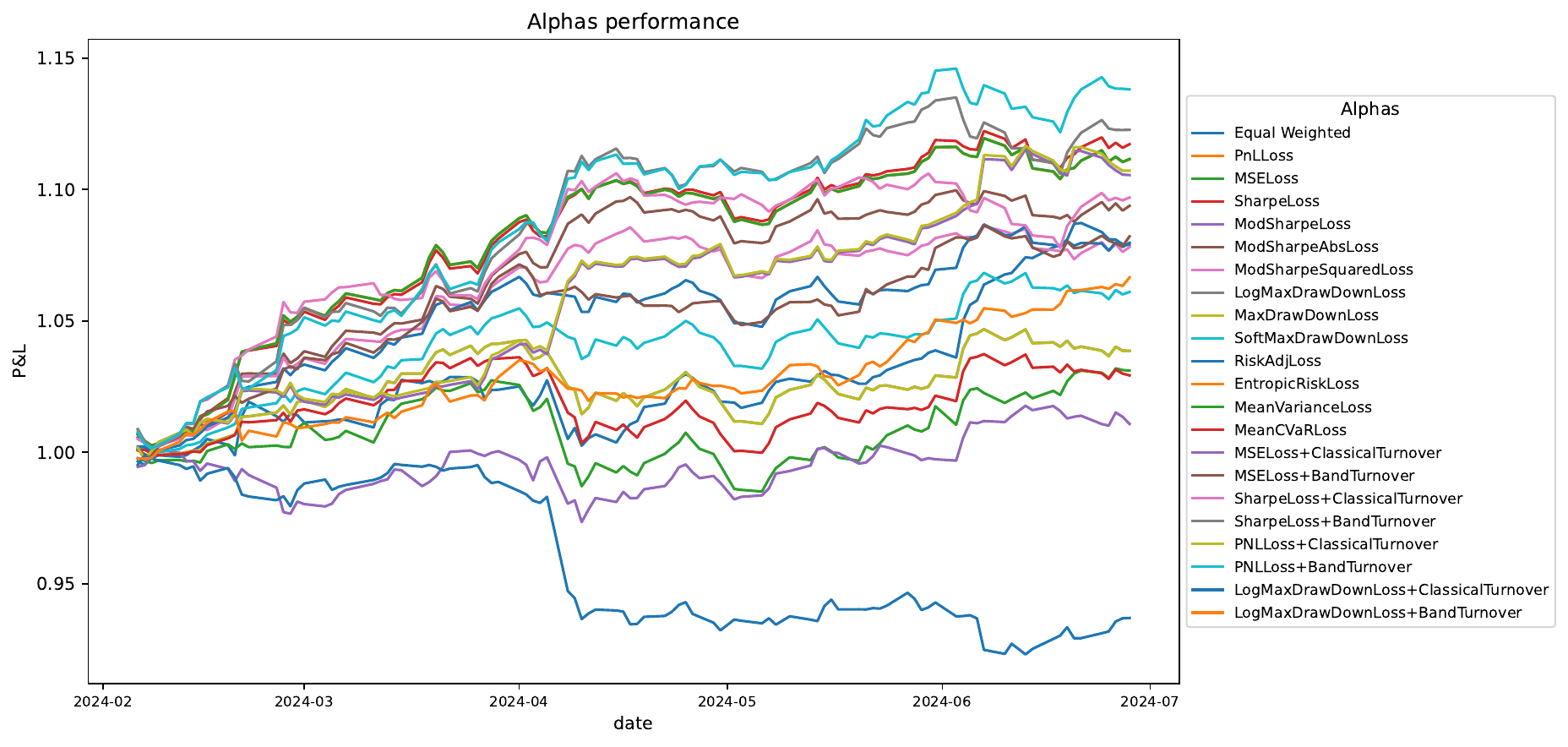} 
    \caption{Alphas correlation heatmap.} 
    \label{fig:portfolio_result} 
\end{figure}

We then investigate different objective functions for learning portfolio-combination weights. The results are summarized in Table~\ref{tab:portfolio_results} and shown in the Figure~\ref{fig:portfolio_result}. Overall, finance-grounded objectives substantially outperform both the equally weighted baseline and classical regression-based weighting schemes, demonstrating the effectiveness of directly optimizing economic performance metrics at the portfolio level.

The strongest performance is achieved by objectives that explicitly control downside risk and incorporate turnover regulation. In particular, the combination of logarithmic maximum drawdown loss with band turnover regularization (\textit{LogMaxDrawDownLoss+BandTurnover}) attains the highest Sharpe ratio while maintaining minimal drawdowns and stable profitability. This highlights the importance of jointly optimizing risk-adjusted returns and trading activity constraints in multi-alpha portfolio construction.

Modified Sharpe-based objectives also perform consistently well, achieving high Sharpe ratios and competitive profit-and-loss outcomes. Compared to the standard \textit{SharpeLoss}, the modified formulations exhibit improved stability and robustness, further validating the proposed scale-sensitive regularization.

In contrast, the equally weighted portfolio and portfolios optimized using standard regression objectives such as \textit{MSELoss} exhibit significantly lower Sharpe ratios and inferior risk-adjusted performance. These results confirm that naive signal aggregation and prediction-error minimization are insufficient for effective portfolio optimization in volatile financial markets.

Overall, the empirical findings demonstrate that combining low-correlated alphas with deep learning–based weighting strategies trained using finance-grounded objectives leads to substantial improvements in portfolio-level performance, particularly when turnover is explicitly controlled within economically meaningful bounds.

\begin{table}[ht!]
\centering
\begin{tabular}{|l|r|r|r|r|}
\hline
\textbf{Alpha} & \textbf{Turnover} & \textbf{Max Drawdown} & \textbf{Profit, \%} & \textbf{Sharpe $\uparrow$} \\\hline
OB\_imbalance\_vol          & 2.199817 & -0.022153 & 0.968738 & 8.889217\\\hline
tr\_val\_bs\_ratio          & 1.304981 & -0.020539 & 0.744540 & 7.986993\\\hline
op\_val\_bs\_ratio          & 2.424682 & -0.027409 & 0.739854 & 6.439757\\\hline
LOB val\_b val\_s ratio     & 0.470399 & -0.059132 & 0.814909 & 5.607485\\\hline
vwap / close                & 5.056355 & -0.114793 & 1.539917 & 5.094586\\\hline
order\_put\_imbalance         & 5.587706 & -0.036191 & 0.711062 & 4.872396\\\hline
reversion                   & 4.287674 & -0.201390 & 1.985513 & 4.752785\\\hline
reverse val\_b / imbalance  & 1.034873 & -0.083021 & 0.623762 & 4.177369\\\hline
high\_low\_time             & 5.270254 & -0.035321 & 0.503467 & 4.012199\\\hline
tpr\_vwap\_ts               & 2.946431 & -0.028602 & 0.433965 & 3.988069\\\hline
put vs cancel ratio         & 0.489127 & -0.055238 & 0.441970 & 3.560326\\\hline
cancel\_vs\_put             & 1.462331 & -0.063046 & 0.650970 & 3.338945\\\hline
lob 10lvl vs bbo spread ratio & 3.860393 & -0.066188 & 0.462961	& 3.072702\\\hline
spread bbo over vwap / close ratio & 2.895454 & -0.086034 & 0.704435 & 3.066959\\\hline
trade\_disb                 & 4.984747 & -0.040891 & 0.303026 & 3.059141\\\hline
vwaps\_ratio                & 5.746206 & -0.066124 & 0.370534 & 2.659815\\\hline
vwap\_1mio\_or\_ratio       & 2.443877 & -0.081554 & 0.534091 & 2.532313\\\hline
cancel\_val / trade\_val    & 0.852424 & -0.043650 & 0.288598 & 2.149364\\\hline
high\_low vwap diff         & 3.124243 & -0.064560 & 0.320390 & 2.103093\\\hline
vol\_imbalance              & 5.461548 & -0.077652 & 0.296717 & 2.048025\\\hline
\end{tabular}
\caption{The table contains evaluation metrics of the classical alphas used to build factor based portfolio. Sorted by Sharpe ratio.}
\label{tab:classical_alphas}
\end{table}

\begin{table}[ht!]
\centering
\small
\begin{tabular}{|l|r|r|r|r|}
\hline
\textbf{Objective} & \textbf{Turnover} & \textbf{Max Drawdown} & \textbf{PnL} & \textbf{Sharpe} \\\hline

LogMaxDrawDownLoss+BandTurnover & 1.0994 & -0.0225 & 0.0666 & \textbf{7.6881} \\\hline
RiskAdjLoss & 0.8362 & -0.0190 & 0.0797 & 7.0307 \\\hline
ModSharpeAbsLoss & 0.7929 & -0.0273 & 0.0938 & 6.3984 \\\hline
ModSharpeSquaredLoss & 0.7638 & -0.0202 & 0.0781 & 6.3968 \\\hline
SharpeLoss+ClassicalTurnover & 0.8272 & -0.0363 & 0.0970 & 6.3156 \\\hline
MSELoss+BandTurnover & 0.8996 & -0.0231 & 0.0822 & 6.1370 \\\hline
SharpeLoss+BandTurnover & 0.7614 & -0.0342 & 0.1227 & 6.0656 \\\hline
SoftMaxDrawDownLoss & 0.8616 & -0.0276 & 0.0610 & 5.8556 \\\hline
SharpeLoss & 0.7378 & -0.0460 & 0.1173 & 5.7360 \\\hline
MeanVarianceLoss & 0.7372 & -0.0501 & 0.1116 & 5.3585 \\\hline
EntropicRiskLoss & 0.7373 & -0.0501 & 0.1116 & 5.3564 \\\hline
PnLLoss & 0.7373 & -0.0502 & 0.1115 & 5.3536 \\\hline
PNLLoss+BandTurnover & 0.7406 & -0.0709 & 0.1381 & 4.0557 \\\hline
LogMaxDrawDownLoss & 0.8608 & -0.0333 & 0.0386 & 3.8489 \\\hline
MaxDrawDownLoss & 0.8607 & -0.0333 & 0.0386 & 3.8478 \\\hline
MeanCVaRLoss & 0.8909 & -0.0375 & 0.0293 & 2.9446 \\\hline
Equal Weighted & 1.2925 & -0.0797 & 0.0788 & 2.5947 \\\hline
PNLLoss+ClassicalTurnover & 0.4310 & -0.1572 & 0.1072 & 1.8283 \\\hline
MSELoss+ClassicalTurnover & 0.4348 & -0.1644 & 0.1055 & 1.6776 \\\hline
MSELoss & 0.9685 & -0.0921 & 0.0311 & 1.0249 \\\hline
LogMaxDrawDownLoss+ClassicalTurnover & 0.6910 & -0.1040 & -0.0631 & -0.1385 \\\hline
ModSharpeLoss & 0.6862 & -0.3403 & 0.0108 & -0.8621 \\\hline

\end{tabular}
\caption{\textbf{Portfolio optimization performance over the test interval, sorted by Sharpe ratio.} 
The table reports turnover, maximum drawdown, cumulative profit-and-loss (PnL), and Sharpe ratio for portfolios constructed using different deep learning–based weighting objectives and the equally weighted baseline. 
Finance-grounded objectives combined with band turnover regularization—particularly \textit{LogMaxDrawDownLoss+BandTurnover} and modified Sharpe-based losses—achieve substantially higher risk-adjusted performance and lower drawdowns compared to classical weighting schemes.}
\label{tab:portfolio_results}
\end{table}

We note that several portfolio optimization objectives achieve unusually high Sharpe ratios, in some cases exceeding values of $7$. Such magnitudes are uncommon in practical trading environments and primarily arise from the high turnover and frequent rebalancing inherent in the evaluated strategies. In the current experimental setup, transaction costs and market impact are not explicitly modeled; consequently, strategies with aggressive trading activity benefit from amplified short-term gains without incurring execution penalties. In realistic market conditions, the inclusion of transaction costs would reduce the effective profitability of high-turnover strategies and lead to substantially lower Sharpe ratios. Nevertheless, the relative performance ranking across objectives remains informative, as all methods are evaluated under identical assumptions. Importantly, objectives combined with band turnover regularization achieve high risk-adjusted returns while maintaining controlled trading activity, suggesting improved robustness under more realistic cost-aware settings.

{\centering
\section{Conclusion and Further Work}}

This work demonstrates that training deep learning models with finance-grounded objectives—such as \textit{ModSharpeLoss}, \textit{SharpeLoss}, \textit{MDDLoss}, and \textit{PnLLoss}—leads to substantially improved economic performance compared to standard regression-based objectives. By directly optimizing risk-adjusted returns and downside risk metrics, the proposed losses produce more robust and profitable algorithmic trading strategies. Furthermore, we show that explicit turnover regularization not only constrains trading activity within economically meaningful bounds but also enhances stability and overall strategy performance.

Several directions for future research naturally follow from these findings. A systematic sensitivity analysis of turnover-regularization hyperparameters would provide deeper insight into the trade-offs between trading activity, transaction efficiency, and risk-adjusted returns. In addition, extending the empirical evaluation across diverse market regimes, asset classes, and longer historical horizons would further assess the generalizability of the proposed framework.

The family of finance-grounded objectives can also be expanded and benchmarked against alternative risk-aware and utility-based loss formulations. Moreover, exploring more advanced deep learning architectures for both alpha generation and portfolio optimization—including xLSTM variants and Transformer-based models—may yield additional performance gains and improved modeling of temporal dependencies.

An especially important extension is the integration of the proposed objectives with limit order book (LOB) data and execution-aware modeling. Incorporating liquidity dynamics and short-term order flow information could enable strategies that explicitly account for market impact and execution constraints, thereby improving realism and robustness in live trading environments.

Finally, the proposed finance-grounded losses naturally translate into reward functions for reinforcement learning frameworks in financial decision-making. Such economically meaningful rewards can produce policies that align closely with practitioner objectives and risk considerations. Embedding these reward structures within language-based agentic systems further opens the possibility of developing interpretable trading agents capable of reasoning about and justifying decisions using quantitative financial criteria, supporting more transparent and trustworthy AI-driven financial systems.

\textbf{Acknowledgments}
The authors thank.....

\textbf{Funding}
\textit{
This work was supported by the grant of the state program of the ``Sirius`` Federal Territory ``Scientific and technological development of the ``Sirius`` Federal Territory`` (Agreement No.~18-03 data 10.09.2024).
}

\textbf{Data availability} The used dataset is published and accessible on Kaggle platform: \url{https://www.kaggle.com/datasets/kkhubiev/cryptotrading}, and the loss function implementation in the jupyter-notebook \url{https://www.kaggle.com/code/kkhubiev/finance-grounded-loss-functions}

\textbf{Ethical Conduct} Not applicable.

\textbf{Conflicts of interest}
The authors declare that there is
no conflict of interest.

\newpage

\appendix
\section{Convergence analysis of gradient optimization for an objective function based on Sharpe's algorithm}\label{app:Analysis}

In this appendix, we 
analyze convergence properties of gradient-based optimization for
the Sharpe-based objective from
Section~\ref{sec:CustomLossFunctions}.

Let $pnl = p_{\boldsymbol{\theta}}$ denote the profit-and-loss random variable induced by the model. In particular, we consider
\[
p_{\boldsymbol{\theta}} = y\, f(\boldsymbol{\theta}, \vec{x}),
\]
where $y$ denotes the realized return and $f(\boldsymbol{\theta}, \vec{x})$ is the model output. Define the mean and variance of $p_{\boldsymbol{\theta}}$ as
\[
\mu(\boldsymbol{\theta}, \vec{x}) = \mathbb{E}\bigl[p_{\boldsymbol{\theta}}\bigr], 
\qquad
V(\boldsymbol{\theta}, \vec{x}) = \mathrm{Var}(p_{\boldsymbol{\theta}}) 
= \mathbb{E}\bigl[p_{\boldsymbol{\theta}}^2\bigr] - \mu(\boldsymbol{\theta}, \vec{x})^2.
\]
Then the Sharpe-based objective in~\eqref{eq:Sharpeloss} can be written as
\begin{equation}
S(\vec\theta, \vec x) = \frac{\mu(\vec\theta, \vec x)}{Var(\vec\theta, \vec x) + \varepsilon}.
\label{eq:sharpe_loss}
\end{equation}

We next derive the gradient of~\eqref{eq:sharpe_loss}. In practice, the expectations in~\eqref{eq:sharpe_loss}--\eqref{eq:sharpe_loss_grad} are approximated using mini-batch estimates. For notational simplicity, let $p=p_{\boldsymbol{\theta}}$ and $f=f(\boldsymbol{\theta}, \vec{x})$. Since $p = y f$, the gradient of $p$ with respect to $\boldsymbol{\theta}$ is

\begin{equation}
    \nabla p = y\nabla f,
\end{equation}
and the gradient of the mean satisfies
\begin{equation}
    \nabla\mu = \nabla\mathbb{E}p=\mathbb{E}\nabla p,
    \label{eq:expetation_grad}
\end{equation}
where we use interchangeability of expectation and differentiation.

Using~\eqref{eq:expetation_grad}, the variance gradient is
\begin{equation}
    \nabla V = \nabla(\mathbb{E}[p^2] -\mu^2) = \mathbb E[2p\nabla p] - 2\mu\nabla\mu =\mathbb{E}[2p\nabla p] - 2\mu\mathbb{E}\nabla p = 2\mathbb{E}[(p - \mu)\nabla p].
    \label{eq:variece_grad}
\end{equation}

Finally, combining~\eqref{eq:expetation_grad} and~\eqref{eq:variece_grad} and substituting into~\eqref{eq:sharpe_loss}, we obtain
\begin{equation}
    \nabla S = \nabla\frac{\mu}{V + \varepsilon} = \frac{(V + \varepsilon)\nabla\mu - \mu\nabla V}{(V+\varepsilon)^2} =\frac{(V + \varepsilon + 2\mu^2)\nabla\mu - 2\mu\mathbb{E}[p\nabla p]}{(V + \varepsilon)^2}.
    \label{eq:sharpe_loss_grad}
\end{equation}

The next step is to analyze convergence properties of gradient-based optimization for the Sharpe-based objective. Let $\mathcal{D}\subset\mathbb{R}^{W}$ be a domain containing the parameter iterates $\{\theta_k\}$ generated by the optimizer. To establish standard convergence guarantees, we seek \emph{Lipschitz smoothness} of the Sharpe objective $S$ in~\eqref{eq:Sharpeloss}, i.e., existence of a constant $L<\infty$ such that
\begin{equation}
    \exists\quad L <\infty\ :\ \mid\mid\nabla S(\theta) - \nabla S(\phi)\mid\mid \leq L\mid\mid \theta -\phi\mid\mid\quad \forall\theta,\phi \in \mathcal{D}.
    \label{eq:sharp_loss_lipsch_req}
\end{equation}

We first note that the denominator in~\eqref{eq:Sharpeloss} is uniformly bounded away from zero over $\mathcal{D}$:
\begin{equation}
    Var(\theta) + \varepsilon \geq\varepsilon \quad \forall\theta\in\mathcal{D},
    \label{eq:variance_nondecreasity}
\end{equation}
which prevents singularities in the objective and its gradient.

We introduce the following boundedness and Lipschitz assumptions on the first-order derivatives of $\mu(\theta)$ and $V(\theta)$ over the domain $\mathcal{D}$. Specifically, let $M_\mu, M_V, L_\mu, L_V, G_\mu,$ and $G_V$ satisfy:
\begin{equation}
    \mid\mid\nabla\mu(\theta) - \nabla\mu(\phi)\mid\mid\leq L_\mu\mid\mid \theta - \phi\mid\mid \quad L_\mu<\infty,
    \label{eq:expectation_lipsch}
\end{equation}
\begin{equation}
    \mid\mid\nabla\mu(\theta)\mid\mid\leq G_\mu <\infty,
    \label{eq:expectation_grad_lipsch}
\end{equation}
\begin{equation}
    \mid\mu(\theta)\mid\leq M_\mu < \infty, \quad \mid V(\theta)\mid \leq M_V<\infty,
    \label{eq:expectation_upper_bound}
\end{equation}
\begin{equation}
    \mid\mid \nabla V(\theta) - \nabla V(\phi)\mid\mid\leq L_V\mid\mid \theta - \phi\mid\mid \quad L_V<\infty,
    \label{eq:variance_lipsch}
\end{equation}
\begin{equation}
    \mid\mid\nabla V(\theta)\mid\mid\leq G_V < \infty.
    \label{eq:variance_grad_lipsch}
\end{equation}

Let $d(\theta)$ be defined as
\begin{equation}
    d(\theta) = Var(\theta) + \varepsilon, \quad \nabla d(\theta)=\nabla V(\theta),
\end{equation}
so that the Sharpe objective gradient can be rewritten as
\begin{equation}
    \nabla S(\theta) = \nabla\frac{\mu(\theta)}{d(\theta)} 
    = \frac{d(\theta)\nabla\mu(\theta) - \mu(\theta)\nabla d(\theta)}{d(\theta)^2} 
    = \frac{\nabla\mu(\theta)}{d(\theta)} -\frac{\mu(\theta)\nabla d(\theta)}{d(\theta)^2}.
    \label{eq:sharpe_loss_grad_mu_d}
\end{equation}
Define
\[
A(\theta) = \frac{\nabla\mu(\theta)}{d(\theta)}, 
\qquad 
B(\theta) = \frac{\mu(\theta)\nabla d(\theta)}{d(\theta)^2}.
\]
Then, by~\eqref{eq:sharp_loss_lipsch_req} and~\eqref{eq:sharpe_loss_grad_mu_d},
\begin{equation}
    \|\nabla S(\theta) - \nabla S(\phi)\| 
    \leq \| A(\theta) - A(\phi)\| + \| B(\theta)- B(\phi)\|
    \leq (L_A + L_B)\|\theta - \phi\|,
\end{equation}
so that $L = L_A + L_B$ serves as a Lipschitz constant for $\nabla S$ on $\mathcal{D}$.

We first bound $\|A(\theta)-A(\phi)\|$. We decompose
\begin{equation}
A(\theta) - A(\phi) 
= \frac{\nabla\mu(\theta)}{d(\theta)} - \frac{\nabla\mu(\phi)}{d(\phi)} 
= \underbrace{\frac{\nabla\mu(\theta)-\nabla\mu(\phi)}{d(\theta)}}_{A_1} 
+\underbrace{\nabla\mu(\phi)\left(\frac{1}{d(\theta)}-\frac{1}{d(\phi)}\right)}_{A_2},
\label{eq:a_diff}
\end{equation}
which implies
\[
\|A(\theta)-A(\phi)\| \leq \|A_1\| + \|A_2\|.
\]
Since $d(\theta) \geq \varepsilon$ for all $\theta\in\mathcal{D}$, we have $\frac{1}{d(\theta)} \leq \frac{1}{\varepsilon}$. Therefore,
\[
\|A_1\| 
= \left\|\frac{1}{d(\theta)}\left[\nabla\mu(\theta)-\nabla\mu(\phi)\right]\right\|
\leq \frac{1}{\varepsilon}\|\nabla\mu(\theta)-\nabla\mu(\phi)\|
\leq \frac{L_\mu}{\varepsilon}\|\theta-\phi\|.
\]
For the second term, note that
\[
\left|\frac{1}{d(\theta)}-\frac{1}{d(\phi)}\right|
= \left|\frac{d(\phi)-d(\theta)}{d(\theta)d(\phi)}\right|
\leq \frac{|d(\phi)-d(\theta)|}{\varepsilon^2}.
\]
Using $|d(\phi)-d(\theta)| = |V(\phi)-V(\theta)|$ and the Lipschitz continuity of $V$ through its gradient bound, we obtain
\[
\|A_2\|
= \left\|\nabla\mu(\phi)\left(\frac{1}{d(\theta)}-\frac{1}{d(\phi)}\right)\right\|
\leq \|\nabla\mu(\phi)\| \cdot \left|\frac{1}{d(\theta)}-\frac{1}{d(\phi)}\right|
\leq G_\mu\frac{G_V}{\varepsilon^2}\|\theta-\phi\|.
\]
Combining the bounds yields
\begin{equation}
    \| A(\theta) - A(\phi)\| 
    \leq \| A_1\| + \| A_2\|
    \leq \frac{1}{\varepsilon^2}\left(\varepsilon L_\mu + G_\mu G_V\right)\|\theta-\phi\|.
    \label{eq:a_norm}
\end{equation}

Secondly, we bound the term $B(\theta)$. We have
\[
B(\theta) - B(\phi) 
= \frac{\mu(\theta)\nabla V(\theta)}{d(\theta)^2} - \frac{\mu(\phi)\nabla V(\phi)}{d(\phi)^2}
= \underbrace{\frac{\mu(\theta)\bigl[\nabla V(\theta) - \nabla V(\phi)\bigr]}{d(\theta)^2}}_{B_1}+\]
\[
+\underbrace{\frac{\bigl[\mu(\theta) - \mu(\phi)\bigr]\nabla V(\phi)}{d(\theta)^2}}_{B_2} - \underbrace{\mu(\phi)\nabla V(\phi)\left(\frac{1}{d(\theta)^2} - \frac{1}{d(\phi)^2}\right)}_{B_3}.
\]
Therefore,
\[
\|B(\theta) - B(\phi)\| \leq \|B_1\| + \|B_2\| + \|B_3\|.
\]

Using~\eqref{eq:expectation_upper_bound},~\eqref{eq:variance_lipsch}, and $d(\theta)^2\geq \varepsilon^2$, we obtain
\[
\|B_1\|
\leq \frac{|\mu(\theta)|}{d(\theta)^2}\,\|\nabla V(\theta)-\nabla V(\phi)\|
\leq \frac{M_\mu}{\varepsilon^2}\,L_V\|\theta-\phi\|.
\]

Using~\eqref{eq:variance_grad_lipsch} and $d(\theta)^2\geq \varepsilon^2$ yields
\[
\|B_2\|
\leq \frac{\|\nabla V(\phi)\|}{d(\theta)^2}\,|\mu(\theta)-\mu(\phi)|
\leq \frac{G_V}{\varepsilon^2}\,|\mu(\theta)-\mu(\phi)|.
\]
Moreover, by boundedness of $\|\nabla\mu(\cdot)\|$ in~\eqref{eq:expectation_grad_lipsch},
\[
|\mu(\theta)-\mu(\phi)| \leq G_\mu \|\theta-\phi\|,
\]
and therefore
\[
\|B_2\| \leq \frac{G_V G_\mu}{\varepsilon^2}\|\theta-\phi\|.
\]

Using~\eqref{eq:expectation_upper_bound} and~\eqref{eq:variance_grad_lipsch}, we have
\[
\|B_3\|
\leq |\mu(\phi)|\,\|\nabla V(\phi)\|\,\left|\frac{1}{d(\theta)^2}-\frac{1}{d(\phi)^2}\right|
\leq M_\mu G_V \left|\frac{d(\phi)^2-d(\theta)^2}{d(\theta)^2d(\phi)^2}\right|.
\]
Since $d(\theta),d(\phi)\geq \varepsilon$,
\[
\|B_3\|
\leq M_\mu G_V\,\frac{|d(\phi)^2-d(\theta)^2|}{\varepsilon^4}
= M_\mu G_V\,\frac{|d(\phi)-d(\theta)|\,|d(\theta)+d(\phi)|}{\varepsilon^4}.
\]
Because $d(\theta)=V(\theta)+\varepsilon \leq M_V+\varepsilon$, it follows that
\[
d(\theta)+d(\phi) \leq 2(M_V+\varepsilon).
\]
In addition,
\[
|d(\theta)-d(\phi)| = |V(\theta)-V(\phi)| \leq G_V\|\theta-\phi\|,
\]
and hence
\[
\|B_3\|
\leq \frac{2M_\mu G_V^2(M_V+\varepsilon)}{\varepsilon^4}\|\theta-\phi\|.
\]

Combining the bounds for $B_1,B_2,$ and $B_3$ yields
\begin{equation}
    \| B(\theta) - B(\phi)\|
    \leq \left(\frac{M_\mu L_V}{\varepsilon^2} + \frac{G_V G_\mu}{\varepsilon^2} + \frac{2M_\mu G_V^2(M_V+\varepsilon)}{\varepsilon^4}\right)\|\theta - \phi\|.
    \label{eq:b_diff}
\end{equation}

Finally, combining~\eqref{eq:a_norm} and~\eqref{eq:b_diff}, we obtain the Lipschitz constant $L$:
\begin{equation}
    L = L_A + L_B 
    = \frac{1}{\varepsilon^2}(\varepsilon L_\mu + G_\mu G_V) 
    + \frac{1}{\varepsilon^4}\left[\varepsilon^2M_\mu L_V + \varepsilon^2G_V G_\mu + 2M_\mu G_V^2(M_V+\varepsilon)\right] < \infty.
\end{equation}

By $L$-smoothness of $S(\theta)$, the standard smoothness (descent) inequality holds:
\begin{equation}
    S(\theta + \Delta) \geq S(\theta) + 
     \langle \nabla S(\theta), \Delta \rangle 
     - \frac{L}{2}\mid\mid\Delta\mid\mid^2,
    \label{eq:ascent_lemma}
\end{equation}
Assume that the optimization iterates satisfy the gradient step
\[
\theta_{k+1} = \theta_k -\eta\nabla S(\theta_k).
\]
Substituting $\Delta = \theta_{k+1}-\theta_k = -\eta\nabla S(\theta_k)$ into~\eqref{eq:ascent_lemma} yields
\[
S(\theta_{k+1})\geq S(\theta_k) - \eta\mid\mid\nabla S(\theta_k)\mid\mid^2 - \frac{L}{2}\eta^2\mid\mid\nabla S(\theta_k)\mid\mid^2,
\]
and therefore
\[
S(\theta_{k+1}) - S(\theta_k)\geq -\eta\left(1 +\frac{L\eta}{2}\right)\mid\mid\nabla S(\theta_k)\mid\mid^2.
\]

To guarantee a monotone decrease of $S(\theta_k)$, it suffices to choose a step size $\eta$ such that $0 < \eta \leq \frac{1}{L}$, which implies $1 + \frac{L\eta}{2}\leq \frac{3}{2}$ and hence
\begin{equation}
    S(\theta_{k+1}) - S(\theta_k)\leq -\frac{\eta}{2}\mid\mid\nabla S(\theta_k)\mid\mid^2 < 0.
\end{equation}
Thus, the sequence $\{S(\theta_k)\}$ is monotone non-increasing. Summing the inequalities from $k=0$ to $K-1$ gives
\[
S(\theta_{K}) - S(\theta_0)\leq - \frac{\eta}{2}\sum_{k=0}^{K-1}\mid\mid\nabla S(\theta_k)\mid\mid^2.
\]

Assuming $S(\theta_k) \geq S_{\min}$ for all $k$, where $S_{\min}$ is a lower bound of the objective on $\mathcal{D}$, we obtain
\begin{equation}
\sum_{k=0}^{\infty}\mid\mid\nabla S(\theta_k)\mid\mid^2 \leq \frac{2(S(\theta_0) - S_{\min})}{\eta} < \infty.
\label{eq:inequalities_sum_up_limit}
\end{equation}

Since the series in~\eqref{eq:inequalities_sum_up_limit} is a sum of nonnegative terms and is finite, it follows that
\[
\mid\mid\nabla S(\theta_k)\mid\mid^2 \rightarrow 0\quad \text{as}\quad k \rightarrow\infty,
\]
otherwise the series would diverge. This establishes first-order stationarity of the iterates, i.e., convergence of the gradient norm to zero.

\newpage
\section{Table of evaluation metrics}
\begin{table}[ht!]
\centering
\scriptsize
\begin{tabular}{|l|r|r|r|r|}
\hline
\textbf{Strategy} & \textbf{Turnover} & \textbf{Max Drawdown} & \textbf{PnL (\%)} & \textbf{Sharpe} \\\hline

LinReg & 1.43 & -0.0870 & 29.64 & 2.0963 \\\hline
LSTM LogMDDLoss & 0.17 & -0.0637 & 16.17 & 1.7573 \\\hline
LSTM MDDLoss & 0.18 & -0.0671 & 15.87 & 1.6914 \\\hline
LSTM ModSharpeAbsLoss+ClassicalTurnover & 0.03 & -0.1453 & 18.72 & 1.5551 \\\hline
LSTM ModSharpeLoss & 0.12 & -0.0670 & 15.28 & 1.5283 \\\hline
LSTM TurnoverLoss+ClassicalTurnover & 0.04 & -0.0405 & 13.42 & 1.4989 \\\hline
LSTM ModSharpeAbsLoss+BandTurnover & 0.04 & -0.1529 & 17.15 & 1.4711 \\\hline
LSTM SharpeLoss & 0.10 & -0.1198 & 12.80 & 1.3628 \\\hline
MLP LogMDDLoss & 0.24 & -0.0919 & 13.78 & 1.2882 \\\hline
LSTM ModSharpeAbsLoss & 0.05 & -0.0978 & 11.02 & 1.2745 \\\hline
LSTM MeanCVaRLoss+BandTurnover & 0.24 & -0.0904 & 15.24 & 1.2579 \\\hline
Reversion & 0.12 & -0.1083 & 24.45 & 1.2418 \\\hline
LSTM SoftMDDLoss & 0.17 & -0.0639 & 11.99 & 1.2158 \\\hline
LSTM ModSharpeLoss+ClassicalTurnover & 0.03 & -0.1789 & 19.42 & 1.0110 \\\hline
LSTM ModSharpeSquaredLoss & 0.07 & -0.1247 & 8.19 & 0.8986 \\\hline
LSTM ModSharpeLoss+BandTurnover & 0.04 & -0.1993 & 17.73 & 0.8959 \\\hline
LSTM TurnoverLoss & 0.04 & -0.0668 & 7.28 & 0.8958 \\\hline
MLP ModSharpeLoss & 0.03 & -0.1905 & 17.15 & 0.8376 \\\hline
LSTM SharpeLoss+ClassicalTurnover & 0.05 & -0.2360 & 12.00 & 0.7764 \\\hline
MLP ModSharpeSquaredLoss & 0.04 & -0.1523 & 11.12 & 0.7732 \\\hline
LSTM SharpeLoss+BandTurnover & 0.06 & -0.2111 & 11.06 & 0.7557 \\\hline
LSTM TurnoverLoss+BandTurnover & 0.04 & -0.0818 & 8.58 & 0.7474 \\\hline
XGBoost & 1.42 & -0.0985 & 25.93 & 0.7181 \\\hline
Mean Reversion & 0.76 & -0.2407 & 14.92 & 0.6480 \\\hline
MLP SoftMDDLoss & 0.26 & -0.0571 & 5.64 & 0.5635 \\\hline
MLP MDDLoss & 0.25 & -0.0829 & 5.00 & 0.4920 \\\hline
LSTM RiskAdjLoss+ClassicalTurnover & 0.03 & -0.2043 & 5.95 & 0.4019 \\\hline
LSTM MeanCVaRLoss & 0.16 & -0.0543 & 3.63 & 0.4014 \\\hline
LSTM SoftMaxDrawDownLoss+ClassicalTurnover & 0.03 & -0.1958 & 5.29 & 0.3428 \\\hline
Buy\&Hold & 0.04 & -0.9226 & 24.90 & 0.3072 \\\hline
LSTM ModSharpeSquaredLoss+BandTurnover & 0.03 & -0.2520 & 5.58 & 0.3045 \\\hline
LSTM MeanVarianceLoss+BandTurnover & 0.04 & -0.2114 & 3.87 & 0.2657 \\\hline
LSTM MeanVarianceLoss+ClassicalTurnover & 0.04 & -0.2170 & 3.74 & 0.2537 \\\hline
LSTM EntropicRiskLoss+ClassicalTurnover & 0.04 & -0.2171 & 3.67 & 0.2482 \\\hline
LSTM EntropicRiskLoss+BandTurnover & 0.04 & -0.2149 & 3.56 & 0.2435 \\\hline
LSTM MeanCVaRLoss+ClassicalTurnover & 0.04 & -0.1389 & 2.26 & 0.2031 \\\hline
MLP ModSharpeAbsLoss & 0.04 & -0.1393 & 2.54 & 0.2007 \\\hline
LSTM DriftAwareTurnoverLoss+ClassicalTurnover & 0.03 & -0.1434 & 2.58 & 0.1862 \\\hline
LSTM ModSharpeSquaredLoss+ClassicalTurnover & 0.03 & -0.2577 & 3.32 & 0.1771 \\\hline
LSTM PnLLoss & 0.12 & -0.1639 & 1.89 & 0.1516 \\\hline
LSTM RiskAdjLoss+BandTurnover & 0.04 & -0.2008 & 1.95 & 0.1370 \\\hline
LSTM PNLLoss+BandTurnover & 0.04 & -0.2410 & 2.14 & 0.1320 \\\hline
MLP SharpeLoss & 0.04 & -0.1647 & 1.46 & 0.1104 \\\hline
LSTM PNLLoss+ClassicalTurnover & 0.04 & -0.2405 & 1.26 & 0.0782 \\\hline
LSTM SoftMaxDrawDownLoss+BandTurnover & 0.23 & -0.2174 & 0.87 & 0.0652 \\\hline
LSTM DriftAwareTurnoverLoss+BandTurnover & 0.03 & -0.1402 & 0.09 & 0.0084 \\\hline
LSTM MeanVarianceLoss & 0.17 & -0.1397 & -0.65 & -0.0614 \\\hline
LSTM LogMaxDrawDownLoss+ClassicalTurnover & 0.04 & -0.1796 & -1.80 & -0.1357 \\\hline
Momentum & 0.49 & -0.1877 & -5.77 & -0.2029 \\\hline
LSTM EntropyLoss & 0.17 & -0.1353 & -4.31 & -0.4135 \\\hline
LSTM LogMaxDrawDownLoss+BandTurnover & 0.24 & -0.1625 & -5.12 & -0.4450 \\\hline
LSTM MSELoss & 0.22 & -0.2004 & -6.03 & -0.4564 \\\hline
LSTM MSELoss+ClassicalTurnover & 0.04 & -0.2408 & -7.43 & -0.5084 \\\hline
MLP MSELoss & 0.59 & -0.1131 & -6.32 & -0.5924 \\\hline
LSTM MaxDrawDownLoss+BandTurnover & 0.27 & -0.1858 & -8.86 & -0.6976 \\\hline
Random Forest & 0.94 & -0.1074 & -7.30 & -0.7808 \\\hline
LSTM MaxDrawDownLoss+ClassicalTurnover & 0.04 & -0.2759 & -18.22 & -1.3424 \\\hline
LSTM MSELoss+BandTurnover & 0.26 & -0.3195 & -20.04 & -1.4305 \\\hline

\end{tabular}
\caption{\textbf{Performance comparison of all evaluated algorithmic trading strategies (alphas) over the test interval, sorted by Sharpe ratio.} 
The table reports portfolio turnover, maximum drawdown, cumulative profit-and-loss (PnL), and Sharpe ratio for heuristic strategies, classical machine learning baselines, and deep learning models trained with different loss functions.}
\label{tab:alphas_result_test_full}
\end{table}

\bibliographystyle{unsrt}

\end{document}